\documentclass{article}

\usepackage{times}
\usepackage{graphicx} 
\usepackage{subfigure} 
\usepackage{geometry,bbm}
\usepackage{comment}
\usepackage{booktabs}
\usepackage{ctable}
\usepackage{pbox}
\usepackage{amssymb}
\usepackage{amsmath}
\usepackage{comment}

\usepackage{natbib}
\usepackage{algorithm}
\usepackage{algorithmic}

\usepackage[accepted,nohyperref]{icml2017}

\icmltitlerunning{Latent Intention Dialogue Models}

\DeclareMathOperator{\E}{\mathbb{E}}

\newcommand*{\Scale}[2][4]{\scalebox{#1}{$#2$}}%

\begin{document} 

\twocolumn[
\icmltitle{Latent Intention Dialogue Models}



\icmlsetsymbol{equal}{*}

\begin{icmlauthorlist}
\icmlauthor{Tsung-Hsien Wen}{cam,equal}
\icmlauthor{Yishu Miao}{ox,equal}
\icmlauthor{Phil Blunsom}{ox}
\icmlauthor{Steve Young}{cam}

\end{icmlauthorlist}

\icmlaffiliation{cam}{Department of Engineering, University of Cambridge, Cambridge, United Kingdom}
\icmlaffiliation{ox}{Department of Computer Science, University of Oxford, Oxford, United Kingdom}

\icmlcorrespondingauthor{Tsung-Hsien Wen}{thw28@cam.ac.uk}
\icmlcorrespondingauthor{Yishu Miao}{yishu.miao@cs.ox.ac.uk}

\icmlkeywords{dialogue, conversation, intention, neural variational inference, latent variable, diverse response, dialogue management}
\vskip 0.3in
]



\printAffiliationsAndNotice{\icmlEqualContribution} 

\begin{abstract}
Developing a dialogue agent that is capable of making autonomous decisions and communicating by natural language is one of the long-term goals of machine learning research. 
Traditional approaches either rely on hand-crafting a small state-action set for applying reinforcement learning that is not scalable or constructing deterministic models for learning dialogue sentences that fail to capture natural conversational variability.
In this paper, we propose a Latent Intention Dialogue Model (LIDM) that employs a discrete latent variable to learn underlying dialogue intentions in the framework of neural variational inference.
In a goal-oriented dialogue scenario, these latent intentions can be interpreted as actions guiding the generation of machine responses, which can be further refined autonomously by reinforcement learning.
The experimental evaluation of LIDM shows that the model out-performs published benchmarks for both corpus-based and human evaluation, demonstrating the effectiveness of discrete latent variable models for learning goal-oriented dialogues. 

\end{abstract}
\section{Introduction}\label{sec:intro}

Recurrent neural networks (RNNs) have shown impressive results in modeling generation tasks that have a sequential structured output form, such as machine translation~\cite{SutskeverVL14,BahdanauCB14}, caption generation~\cite{KarpathyF14,xu2015icml}, and natural language generation~\cite{wensclstm15,Kiddon2016}. 
These discriminative models are trained  to learn only a conditional output distribution over strings and despite the sophisticated architectures and conditioning mechanisms used to ensure salience, they are not able to model the underlying actions needed to generate natural dialogues.
As a consequence,  these sequence-to-sequence models are limited in their ability to exhibit the intrinsic variability and stochasticity of natural dialogue.
For example  both goal-oriented dialogue systems~\cite{wenN2N17,bordes16n2n} and sequence-to-sequence learning chatbots~\cite{VinyalsL15,ShangLL15,SerbanSBCP15} struggle to generate  diverse yet causal responses~\cite{LiGBGD15,serban2016lv}.
In addition, there is often insufficient training data for goal-oriented dialogues which results in over-fitting which prevents  deterministic models from learning effective and scalable interactions.
In this paper, we propose a latent variable model -- Latent Intention Dialogue Model (LIDM) -- for learning the complex distribution of communicative intentions in goal-oriented dialogues. 
Here, the latent variable representing dialogue intention can be considered as the autonomous decision-making center of a dialogue agent for composing appropriate machine responses.


Recent advances in neural variational inference ~\cite{icml2014Kingma,mnih14NVIL} have sparked a series of latent variable models applied to NLP ~\cite{BowmanVVDJB15,serban2016lv,miaoTVAE16,cao17eacl}.
For models with continuous latent variables, the reparameterisation trick~\cite{icml2014Kingma} is commonly used to build an unbiased and low-variance gradient estimator for updating the models.
However, since a continuous latent space is hard to interpret, the major benefits of these models are the stochasticity and the regularisation brought by the latent variable.
In contrast, models with discrete latent variables are able to not only produce interpretable latent distributions but also provide a principled framework for semi-supervised learning~\cite{NIPS2014_5352}.
This is critical for  NLP tasks, especially where additional supervision and external knowledge can be utilized for bootstrapping~\cite{FaruquiDJDHS15,miao16latentLangauage,kocisky16}.
However, variational inference with discrete latent variables is relatively difficult due to the problem of high variance during sampling.
Hence
we introduce baselines, as in the REINFORCE~\cite{Williams1992} algorithm, to mitigate the high variance problem, and carry out efficient neural variational inference~\cite{mnih14NVIL} for the latent variable model.

In the LIDM, the latent intention is inferred from user input utterances. 
Based on the dialogue context, the agent draws a sample as the intention which then guides the natural language response generation.
Firstly, in the framework of neural variational inference~\cite{mnih14NVIL}, we construct an inference network to approximate the posterior distribution over the latent intention.
Then, by sampling the intentions for each response, we are able to directly learn a basic intention distribution on a human-human dialogue corpus by optimising the variational lower bound.
To further reduce the variance, we utilize a labeled subset of the corpus in which the labels of intentions are automatically generated by clustering.
Then, the latent intention distribution can be learned in a semi-supervised  fashion, where the learning signals are either from the direct supervision (labeled set) or the variational lower bound (unlabeled set).

From the perspective of reinforcement learning, the latent intention distribution can be interpreted as the intrinsic policy that reflects human decision-making under a particular conversational scenario. 
Based on the initial policy (latent intention distribution) learnt from the semi-supervised variational inference framework, the model can refine its strategy easily against alternative objectives using policy gradient-based reinforcement learning.
This is somewhat analagous to the training process used in AlphaGo \cite{silver2016mastering} for the game of Go.
Based on LIDM, we show that different learning paradigms can be brought together under the same framework to bootstrap the development of a dialogue agent~\cite{liEMNLP20162,NIPS2016_6264}.

In summary, the contribution of this paper is two-fold: firstly, we show that the neural variational inference framework is able to discover discrete, interpretable intentions from data to form the decision-making basis of a dialogue agent;
secondly, the agent is capable of revising its conversational strategy based on an external reward within the same framework.
This is important because it provides a stepping stone towards building an autonomous dialogue agent that can continuously improve itself through interaction with users.  
The experimental results demonstrate the effectiveness of our latent intention model which achieves  state-of-the-art performance on both automatic corpus-based evaluation and human evaluation.


\section{Latent Intention Dialogue Model for Goal-oriented Dialogue}\label{sec:LIDM}

Goal-oriented dialogue\footnote{Like most of the goal-oriented dialogue research, we focus on information seek type dialogues.}~\cite{6407655} aims at building models that can help users to complete certain tasks via natural language interaction.
Given a user input utterance $u_t$ at turn $t$ and a knowledge base (KB), the model needs to parse the input into actionable commands $Q$ and access the KB to search for useful information in order to answer the query.
Based on the search result, the model needs to summarise its findings and reply with an appropriate response $m_t$ in natural language.

\begin{figure*}[t]
\centerline{\includegraphics[width=160mm]{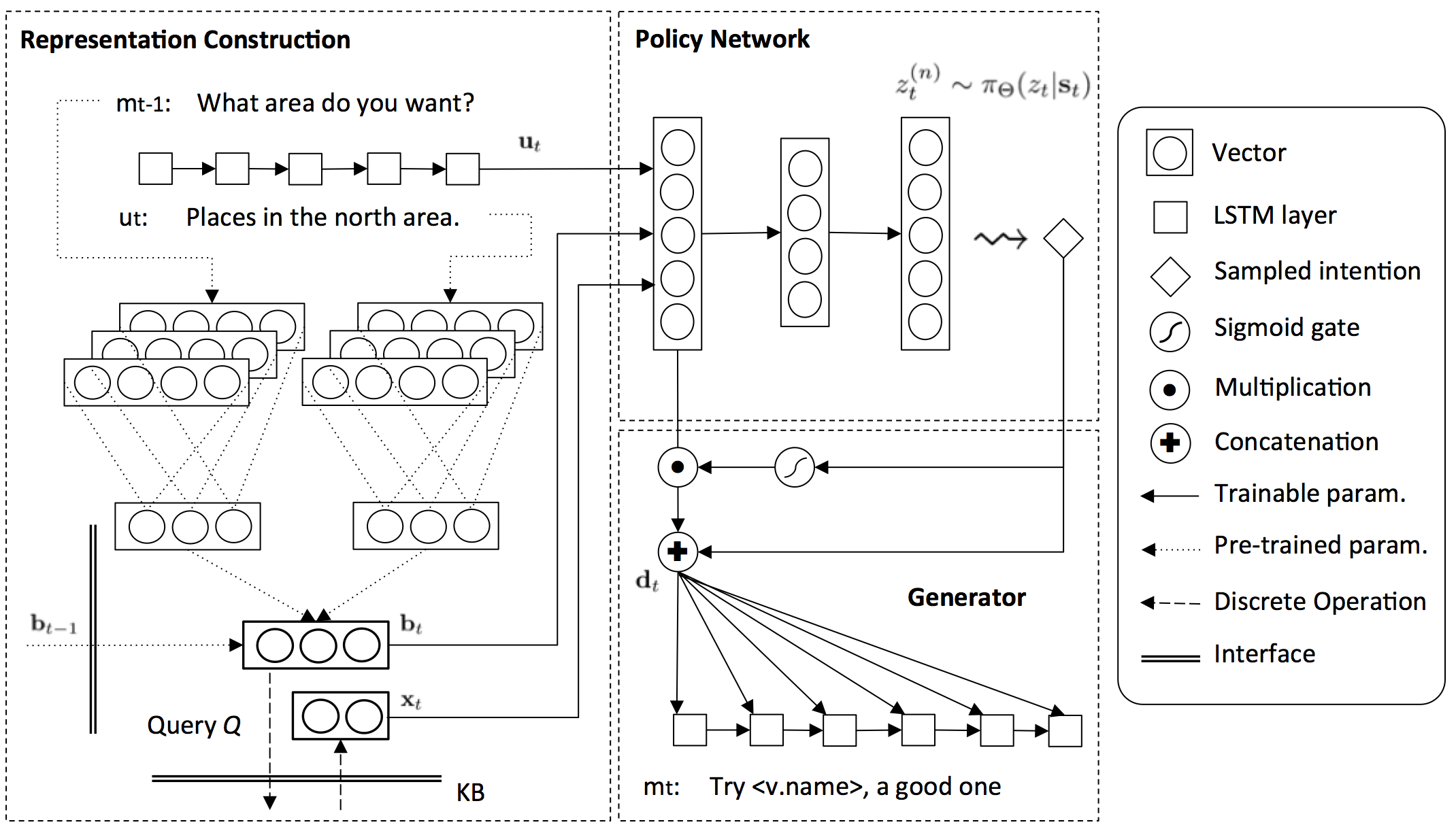}}
\caption{LIDM for Goal-oriented Dialogue Modeling}
\label{fig:lidm}
\end{figure*}

\subsection{Model}\label{ssec:lidm_model}

The LIDM is based on the end-to-end  system architecture described in~\cite{wenN2N17}.  It comprises three components: (1) Representation Construction; (2) Policy Network; and (3) Generator, as shown in Figure~\ref{fig:lidm}.
To capture the user's intent and match it against the system's knowledge, a dialogue state vector $\mathrm{\mathbf{s}}_t =\mathrm{\mathbf{u}}_t \oplus \mathrm{\mathbf{b}}_t \oplus \mathrm{\mathbf{x}}_t$ is derived from the user input $u_t$ and the knowledge base KB: 
$\mathrm{\mathbf{u}}_t$ is the distributed utterance representation, which is formed by encoding the user utterance\footnote{\label{fn:delex}All sentences are pre-processed by delexicalisation~\cite{henderson14} where slot-value specific words are replaced with their corresponding generic tokens based on an ontology.} $u_t$ with a bidirectional LSTM~\cite{Hochreiter1997} and concatenating the final stage hidden states together, 
\begin{equation}\label{eq:bilstm}
\mathrm{\mathbf{u}}_t = \text{biLSTM}_\Theta(u_t). 
\end{equation}
The belief vector $\mathrm{\mathbf{b}}_t$, which is a concatenation of a set of probability distributions over domain specific slot-value pairs, is extracted by a set of pre-trained RNN-CNN belief trackers~\cite{wenN2N17,mrksic2016nbt}, in which $u_t$ and $m_{t-1}$ are processed by two different CNNs as shown in Figure~\ref{fig:lidm},
\begin{equation}
\mathrm{\mathbf{b}}_t = \text{RNN-CNN}(u_t, m_{t-1}, \mathrm{\mathbf{b}}_{t-1})
\end{equation}
where $m_{t-1}$ is the preceding machine response and $\mathrm{\mathbf{b}}_{t-1}$ is the preceding belief vector.
They are included to model the current turn of the discourse and the long-term dialogue context, respectively.
Based on the belief vector, a query $Q$ is formed by taking the union of the maximum values of each slot. 
$Q$ is then used to search the internal KB and return a vector $\mathrm{\mathbf{x}}_{t}$ representing the degree of matching in the KB.
This is produced by counting all the matching venues and re-structuring it into a six-bin one-hot vector. 
Among the three vectors that comprise the dialogue state $\mathrm{\mathbf{s}}_t$, $\mathrm{\mathbf{u}}_t$ is completely trainable from data, $\mathrm{\mathbf{b}}_t$ is pre-trained using a separate objective function, and $\mathrm{\mathbf{x}}_t$ is produced by a discrete database accessing operation.
For more details about the belief trackers and database operation  refer to Wen et al~\yrcite{wencond16,wenN2N17}.

Conditioning on the state $\mathrm{\mathbf{s}}_{t}$, the policy network parameterises the latent intention $z_t$ by a single layer MLP,
\begin{equation}\label{eq:pi}
\pi_\Theta(z_t|\mathrm{\mathbf{s}}_{t}) = \text{softmax}(\mathrm{\mathbf{W}}_{2}^\intercal\cdot\tanh(\mathrm{\mathbf{W}}_{1}^\intercal\mathrm{\mathbf{s}}_{t}+\mathrm{\mathbf{b}}_{1})+\mathrm{\mathbf{b}}_{2})
\end{equation}
where $\mathrm{\mathbf{W}}_{1}$, $\mathrm{\mathbf{b}}_{1}$, $\mathrm{\mathbf{W}}_{2}$, $\mathrm{\mathbf{b}}_{2}$ are model parameters.
Since $\pi_\Theta(z_t|\mathrm{\mathbf{s}}_{t})$ is a discrete conditional probability distribution based on dialogue state, we can also interpret the policy network here as a latent dialogue management component in the traditional POMDP-based framework~\cite{6407655,6639297}.
A latent intention $z_t^{(n)}$ (or an action in the reinforcement learning literature)  can then be sampled from the conditional distribution,
\begin{equation}\label{eq:sample}
z_t^{(n)} \sim \pi_\Theta(z_t|\mathrm{\mathbf{s}}_{t}).
\end{equation}
This sampled intention (or action) $z_t^{(n)}$ and the state vector $\mathrm{\mathbf{s}}_{t}$ can then be combined into a control vector $\mathrm{\mathbf{d}}_{t}$, which is used to govern the generation of the system response based on a conditional LSTM language model,
\begin{equation}\label{eq:dt}
\mathrm{\mathbf{d}}_{t} = \mathrm{\mathbf{W}}_{4}^\intercal\mathrm{\mathbf{z}}_{t} \oplus 
\left[\text{sigmoid}(\mathrm{\mathbf{W}}_{3}^\intercal\mathrm{\mathbf{z}}_{t}+\mathrm{\mathbf{b}}_{3})\cdot \mathrm{\mathbf{W}}_{5}^\intercal\mathrm{\mathbf{s}}_{t}\right]
\end{equation}
\begin{equation}\label{eq:obj}
p_\Theta(m_t|\mathrm{\mathbf{s}}_{t},z_t) = \prod_j p(w_{j+1}^{t}|w_{j}^{t},\mathrm{\mathbf{h}}_{j-1}^{t},\mathrm{\mathbf{d}}_{t})
\end{equation}
where $\mathrm{\mathbf{b}}_{3}$ and $\mathrm{\mathbf{W}}_{3\sim 5}$ are parameters, $\mathrm{\mathbf{z}}_{t}$ is the 1-hot representation of $z_t^{(n)}$, $w_{j}^{t}$ is the last output token (i.e. a word, a delexicalised\textsuperscript{\ref{fn:delex}} slot name or a delexicalised\textsuperscript{\ref{fn:delex}} slot value), and $\mathrm{\mathbf{h}}_{j-1}^t$ is the decoder's last hidden state.
Note in Equation~\ref{eq:dt} the degree of information flow from the state vector is controlled by a sigmoid gate whose input signal is the sampled intention $z_t^{(n)}$. 
This  prevents the decoder from over-fitting to the deterministic state information and forces it to take the sampled stochastic intention into account.
The LIDM can then be formally written down in its parameterised form with  parameter set $\Theta$,
\begin{equation}\label{eq:lidm_model} p_\Theta(m_t|\mathrm{\mathbf{s}}_{t})=\sum_{z_t}p_\Theta(m_t|z_t,\mathrm{\mathbf{s}}_{t})\pi_\Theta(z_t|\mathrm{\mathbf{s}}_{t}).
\end{equation}

\subsection{Inference}\label{ssec:inf_lidm}

To carry out inference for the LIDM, we introduce an inference network $q_\Phi(z_t|\mathrm{\mathbf{s}}_{t},m_t)$ to approximate the posterior distribution $p(z_t|\mathrm{\mathbf{s}}_{t},m_t)$ so that we can optimise the variational lower bound of the joint probability in a neural variational inference framework \cite{miaoTVAE16}. 
We can then derive the variational lower bound as,
\begin{align}\label{eq:vlb}
\mathcal{L} &= \Scale[0.95]{\E_{q_\Phi(z_t)} [\log p_\Theta(m_t|z_t,\mathrm{\mathbf{s}}_{t})] - \lambda D_{KL}(q_\Phi(z_t)||\pi_{\Theta}(z_t|\mathrm{\mathbf{s}}_{t}))}\nonumber\\
&\leq \log \sum_{z_t} p_\Theta(m_t|z_t,\mathrm{\mathbf{s}}_{t}) \pi_{\Theta}(z_t|\mathrm{\mathbf{s}}_{t}) \nonumber\\
&=\log p_\Theta(m_t|\mathrm{\mathbf{s}}_{t})
\end{align}
where $q_\Phi(z_t)$ is a shorthand for $q_\Phi(z_t|\mathrm{\mathbf{s}}_{t},m_t)$.
Note that we use a modified version of the lower bound here by incorporating a trade-off factor $\lambda$~\cite{betaVAE17}.
The inference network $q_\Phi(z_t|\mathrm{\mathbf{s}}_{t},m_t)$ is then constructed by
\begin{equation}\label{eq:q}
q_\Phi(z_t|\mathrm{\mathbf{s}}_{t},m_t) = \text{Multi}(\mathrm{\mathbf{o}}_{t})=\text{softmax}(\mathrm{\mathbf{W}}_{6}\mathrm{\mathbf{o}}_{t})
\end{equation}
\begin{equation}\label{eq:joint}
\mathrm{\mathbf{o}}_{t} = \text{MLP}_\Phi(\mathrm{\mathbf{b}}_{t},\mathrm{\mathbf{x}}_{t},\mathrm{\mathbf{u}}_{t},\mathrm{\mathbf{m}}_{t})
\end{equation}
\begin{equation}\label{eq:infstate}
\mathrm{\mathbf{u}}_{t} = \text{biLSTM}_\Phi(u_t), \mathrm{\mathbf{m}}_{t} = \text{biLSTM}_\Phi(m_t)
\end{equation}
where $\mathrm{\mathbf{o}}_{t}$ is the joint representation, and both $\mathrm{\mathbf{u}}_{t}$ and $\mathrm{\mathbf{m}}_{t}$ are modeled by a bidirectional LSTM network.
Although both $q_\Phi(z_t|\mathrm{\mathbf{s}}_{t},m_t)$ and $\pi_{\Theta}(z_t|\mathrm{\mathbf{s}}_{t})$ are modelled as parameterised multinomial distributions, the approximation $q_\Phi(z_t|\mathrm{\mathbf{s}}_{t},m_t)$  only functions during inference by producing samples to compute the stochastic gradients, while $\pi_{\Theta}(z_t|\mathrm{\mathbf{s}}_{t})$ is the generative distribution that generates the required samples for composing the machine response.

Based on the samples $z_t^{(n)} \sim q_\Phi(z_t|\mathrm{\mathbf{s}}_{t},m_t)$, we use different strategies to alternately optimise the parameters $\Theta$ and $\Phi$ against the variational lower bound (Equation~\ref{eq:vlb}).
To do this, we  further divide $\Theta$ into two sets  $\Theta=\{\Theta_1,\Theta_2\}$.
Parameters $\Theta_1$ on the decoder side are directly updated  by back-propagating the gradients,
\begin{align}\label{eq:decgrad}
\frac{\partial\mathcal{L}}{\partial \Theta_1} &= \E_{q_\Phi(z_t|\mathrm{\mathbf{s}}_{t},m_t)} [
\frac{\partial\log p_{\Theta_1}(m_t|z_t,\mathrm{\mathbf{s}}_{t})}{\partial \Theta_1}]\nonumber\\
&~\approx \frac{1}{N}\sum_n\frac{\partial\log p_{\Theta_1}(m_t|z_t^{(n)},\mathrm{\mathbf{s}}_{t})}{\partial \Theta_1}.
\end{align}
Parameters $\Theta_2$ in the generative network, however,  are updated by minimising the KL divergence,
\begin{align}\label{eq:klgrad}
\frac{\partial\mathcal{L}}{\partial \Theta_2} &= -\frac{\partial \lambda D_{KL}(q_\Phi(z_t|\mathrm{\mathbf{s}}_t,m_t)||\pi_{\Theta_2}(z_t|\mathrm{\mathbf{s}}_{t}))}{\partial \Theta_2}\nonumber\\
&= -\lambda\sum_{z_t}{q_\Phi(z_t|\mathrm{\mathbf{s}}_t,m_t)} \frac{\partial  \log \pi_{\Theta_2}(z_t|\mathrm{\mathbf{s}}_{t})}{\partial\Theta_2}
\end{align}
where the entropy derivative $\partial H[q_\Phi(z_t|\mathrm{\mathbf{s}}_t,m_t)]/\partial \Theta_2 = 0$ and therefore can be ignored.
Finally, for the parameters $\Phi$ in the inference network, we firstly define the learning signal $r(m_t,z_t^{(n)},\mathrm{\mathbf{s}}_{t})$,
\begin{align}\label{eq:ls}
r(m_t,&z_t^{(n)},\mathrm{\mathbf{s}}_{t}) = \log p_{\Theta_1}(m_t|z_t^{(n)},\mathrm{\mathbf{s}}_{t}) - \nonumber\\
&\lambda (\log q_\Phi(z_t^{(n)}|\mathrm{\mathbf{s}}_t,m_t) - \log \pi_{\Theta_2}(z_t^{(n)}|\mathrm{\mathbf{s}}_{t})).
\end{align}
Then the parameters $\Phi$ are updated by,
\begin{align}\label{eq:phigrad}
\frac{\partial\mathcal{L}}{\partial \Phi} 
&= \E_{q_\Phi(a_t|\mathrm{\mathbf{s}}_{t},m_t)} [r(m_t,a_t,\mathrm{\mathbf{s}}_{t}) \frac{\partial \log q_\Phi(a_t|\mathrm{\mathbf{s}}_{t},m_t)}{\partial \Phi} ]\nonumber\\
&\approx \frac{1}{N} \sum_{n} r(m_t,z_t^{(n)},\mathrm{\mathbf{s}}_{t}) \frac{\partial \log q_\Phi(z_t^{(n)}|\mathrm{\mathbf{s}}_{t},m_t)}{\partial \Phi}.
\end{align}
However, this gradient estimator has a large variance because the learning signal $r(m_t,z_t^{(n)},\mathrm{\mathbf{s}}_{t})$ relies on samples from the proposal distribution $q_\Phi(z_t|\mathrm{\mathbf{s}}_{t},m_t)$.
To reduce the variance during inference, we follow the REINFORCE algorithm~\cite{NIPS2014_5542,mnih14NVIL} and introduce two baselines $b$ and $b(\mathrm{\mathbf{s}}_{t})$, the centered learning signal and input dependent baseline respectively to help reduce the variance.
$b$ is a learnable constant and $b(\mathrm{\mathbf{s}}_{t})=\text{MLP}(\mathrm{\mathbf{s}}_{t})$.
During training, the two baselines are updated by minimising the distance,
\begin{equation}\label{eq:bsobj}
\mathcal{L}_b = \left[r(m_t,z_t^{(n)},\mathrm{\mathbf{s}}_{t})-b-b(\mathrm{\mathbf{s}}_{t})\right]^2
\end{equation}
and the gradient w.r.t. $\Phi$ can be rewritten as
\begin{equation}\label{eq:phigrad2}
\frac{\partial\mathcal{L}}{\partial \Phi} 
\approx \frac{1}{N}\sum_{n} [r(m_t,z_t^{(n)},\mathrm{\mathbf{s}}_{t})-b-b(\mathrm{\mathbf{s}}_{t})] \frac{\partial \log q_\Phi(z_t^{(n)}|\mathrm{\mathbf{s}}_{t},m_t)}{\partial \Phi}.
\end{equation}

\vspace{-3mm}
\subsection{Semi-Supervision}\label{ssec:sssl}

Despite the steps described above for reducing the variance, there remain two major difficulties in learning latent intentions in a completely unsupervised manner: 
(1) the high variance of the inference network prevents it from generating sensible intention samples in the early stages of training, and 
(2) the overly strong discriminative power of the LSTM language model is prone to the {\it disconnection} phenomenon between the LSTM decoder and the rest of the components whereby the decoder learns to ignore the samples and focuses solely on optimising the language model.
To ensure more stable training and prevent disconnection, a semi-supervised learning technique is introduced.

Inferring the latent intentions underlying utterances is similar to an unsupervised clustering task.
Standard clustering algorithms can therefore be used to pre-process the corpus and generate automatic labels $\hat{z_t}$ for part of the training examples $(m_t,\mathrm{\mathbf{s}}_{t},\hat{z_t}) \in \mathbb{L}$.
Then when the model is trained on the unlabeled examples $(m_t,\mathrm{\mathbf{s}}_{t})\in \mathbb{U}$, we optimise it against the modified variational lower bound given in Equation~\ref{eq:vlb}
\begin{align}\label{eq:l1}
\mathcal{L}_1 = \sum_{(m_t,\mathrm{\mathbf{s}}_{t})\in \mathbb{U}}
&\E_{q_\phi(z_t|\mathrm{\mathbf{s}}_{t},m_t)} \left[\log p_\theta(m_t|z_t,\mathrm{\mathbf{s}}_{t})\right] \nonumber\\
&- \lambda D_{KL}(q_\phi(z_t|\mathrm{\mathbf{s}}_{t},m_t)||\pi_{\theta}(z_t|\mathrm{\mathbf{s}}_{t}))
\end{align}
However, when the model is updated based on examples from the labeled set $(m_t,\mathrm{\mathbf{s}}_{t},\hat{z_t}) \in \mathbb{L}$, we treat the labeled intention $\hat{z_t}$ as an observed variable and train the model by maximising the joint log-likelihood,
\begin{equation}\label{eq:l2}
\mathcal{L}_2 = 
\sum_{(m_t,\hat{z_t},\mathrm{\mathbf{s}}_{t})\in \mathbb{L}} \log\left[ p_\Theta(m_t|\hat{z_t},\mathrm{\mathbf{s}}_{t}) \pi_\Theta(\hat{z_t}|\mathrm{\mathbf{s}}_{t})q_\Phi(\hat{z_t}|\mathrm{\mathbf{s}}_{t},m_t)\right]
\end{equation}
The final joint objective function can then be written as $\mathcal{L'}=\alpha\mathcal{L}_1+\mathcal{L}_2$, where $\alpha$ 
controls the trade-off between the supervised and unsupervised examples.

\subsection{Reinforcement Learning}\label{ssec:rl}

One of the main purposes of learning interpretable, discrete latent intention inside a dialogue system is to be able to control and refine the model's behaviour with operational experience.
The learnt generative network $\pi_\Theta(z_t|\mathrm{\mathbf{s}}_{t})$ encodes the policy discovered from the underlying data distribution but this is not necessarily optimal for any specific task.
Since $\pi_\Theta(z_t|\mathrm{\mathbf{s}}_{t})$ is a parameterised policy network itself, any policy gradient-based reinforcement learning algorithm~\cite{Williams1992,Konda2003} can be used to fine-tune the initial policy against other objective functions that we are more interested in.

Based on the initial policy $\pi_\Theta(z_t|\mathrm{\mathbf{s}}_{t})$, we revisit the training dialogues and update parameters based on the following strategy: 
when encountering unlabeled examples $\mathbb{U}$ at turn $t$ the system samples an action from the learnt policy $z_t^{(n)} \sim \pi_\Theta(z_t|\mathrm{\mathbf{s}}_{t})$ and receives a reward $r_t^{(n)}$.
Conditioning on these, we can directly fine-tune a subset of the model parameters $\Theta'$ by the policy gradient method,
\begin{align}\label{eq:rlobj}
\frac{\partial\mathcal{J}}{\partial\Theta'} 
&\approx \frac{1}{N}\sum_{n} r_t^{(n)} \frac{\partial \log \pi_\Theta(z_t^{(n)}|\mathrm{\mathbf{s}}_{t})}{\partial\Theta'}
\end{align}
where $\Theta'=\{\mathrm{\mathbf{W}}_{1},\mathrm{\mathbf{b}}_{1},\mathrm{\mathbf{W}}_{2},\mathrm{\mathbf{b}}_{2}\}$ is the MLP that parameterises the policy network (Equation~\ref{eq:pi}).
However, when a labeled example $\in \mathbb{L}$ is encountered we force the model to take the labeled action $z_t^{(n)}=\hat{z_t}$ and update the parameters by Equation~\ref{eq:rlobj} as well.
Unlike Li et al~\yrcite{liEMNLP20162} where the whole model is refined end-to-end using RL, updating only $\Theta'$ effectively allows us to refine only the {\it decision-making} of the system and avoid the problem of over-fitting.

\section{Experiments}\label{sec:exp}

\subsection{Dataset \& Setup for Goal-oriented Dialogue}\label{ssec:setupGOD}

We explored the properties of the LIDM model\footnote{Will be available at https://github.com/shawnwun/NNDIAL} using the CamRest676 corpus\footnote{https://www.repository.cam.ac.uk/handle/1810/260970} collected by  Wen et al~\yrcite{wenN2N17}, in which the task of the system is to assist users to find a restaurant in the Cambridge, UK area.
The corpus was collected based on a modified Wizard of Oz~\cite{Kelley84} online data collection.
Workers were recruited on Amazon Mechanical Turk and asked to complete a task by carrying out a conversation, alternating roles between a user and a wizard.
There are three informable slots ({\it food}, {\it pricerange}, {\it area}) that users can use to constrain the search and six requestable slots ({\it address}, {\it phone}, {\it postcode} plus the three informable slots) that the user can ask a value for once a restaurant has been offered.
There are 676 dialogues in the dataset (including both finished and unfinished dialogues) and approximately 2750 conversational turns in total.
The database contains 99 unique restaurants.

To make a direct comparison with  prior work we follow the same experimental setup as in Wen et al~\yrcite{wencond16,wenN2N17}.
The corpus was partitioned into training, validation, and test sets in the ratio 3:1:1. 
The LSTM hidden layer sizes were set to 50, and the vocabulary size is around 500 after pre-processing, 
to remove rare words and words that can be delexicalised\textsuperscript{\ref{fn:delex}}.
All the system components were trained jointly by fixing the pre-trained belief trackers and the discrete database operator with the model's latent intention size $I$ set to 50, 70, and 100, respectively.
The trade-off constants $\lambda$ and $\alpha$ were both set to 0.1.
To produce self-labeled response clusters for semi-supervised learning of the intentions, we firstly removed function words from all the responses and clustered them according to their content words.
We then assigned the responses in the $i$-th frequent cluster to the $i$-th latent dimension as its supervised set.
This results in about 35\% ($I=50$) to 43\% ($I=100$) labeled responses across the whole dataset.
An example of the resulting seed set is shown in Table~\ref{tab:seed}.
During inference we carried out stochastic estimation by taking one sample for estimating the stochastic gradients.
The model is trained by Adam~\cite{KingmaB14} and tuned (early stopping, hyper-parameters) on the held-out validation set.
We alternately optimised the generative model and the inference network by fixing the parameters of one while updating the parameters of the other.

\begin{table}[t]
  	\centering
  	\Scale[1]{
  	\begin{tabular}{lll}
    \toprule
    ID	&	\#	&	content words\\
    \midrule
  	0	&	138	&	thank, goodbye\\
    1	&	91	&	welcome, goodbye\\
    3	&	42	&	phone, address, [v.phone],[v.address]\\
    14	&	17	&	address, [v.address]\\
    31	&	9	&	located, area, [v.area]\\
    34	&	9	&	area, would, like\\
    46	&	7	&	food, serving, restaurant, [v.food]\\
    85	&	4	& 	help, anything, else\\
    \bottomrule
	\end{tabular}}
    \caption{An example of the automatically labeled response seed set for semi-supervised learning during variational inference.}
    \vspace{-3mm}
    \label{tab:seed}
\end{table}

During reinforcement fine-tuning, we generated a sentence $m_t$ from the model to replace the ground truth $\hat{m_t}$ at each turn and define an immediate reward as whether $m_t$ can improve the dialogue success~\cite{SuVGKMWY15} relative to
$\hat{m_t}$, plus the sentence BLEU score~\cite{export217163},
\begin{equation}\label{eq:reward}
\Scale[0.95]{
r_t = \eta\cdot\text{sBLEU}(m_t,\hat{m_t})+
\begin{cases}
      1 & m_t \text{ improves }\\
      -1& m_t \text{ degrades }\\
      0&  \text{otherwise}
   \end{cases}}
\end{equation}
where the constant $\eta$ was set to 0.5.
We fine-tuned the model parameters using RL for only 3 epochs.
During testing, we greedily selected the most probable intention and applied beam search with the beamwidth set to 10 when decoding the response.
The decoding criterion was the average log-probability of tokens in the response.
We then evaluated our model on task success rate~\cite{SuVGKMWY15} and BLEU score~\cite{papineni2002bleu} as in Wen et al~\yrcite{wencond16,wenN2N17} in which the model is used to predict each system response in the held-out test set.

\subsection{Experiments on Goal-oriented Dialogue}\label{ssec:expGOD}

\begin{table}[t]
  \centering
  \Scale[1]{\begin{tabular}{llcc}
    \toprule
    &		Model 		&	Success (\%)	&	BLEU\\
    \midrule
    \multicolumn{4}{l}{\bf Ground Truth}\\
    &	Ground Truth	&	91.6	&	1.000\\
    \midrule
    \multicolumn{4}{l}{\bf Published Models~\cite{wencond16}}\\
    &	NDM			&  	76.1	& 	0.212\\
    &	NDM+Att		& 	79.0	& 	0.224\\
    &	NDM+Att+SS	& 	81.8	& 	0.240\\
    \midrule
    \multicolumn{4}{l}{\bf LIDM Models}\\
    &	LIDM, $I=50$		&	66.9	&	0.238\\
    & 	LIDM, $I=70$		&	61.0	&	{\bf0.246}\\
    &	LIDM, $I=100$		&	63.2	&	0.242\\
    \midrule
	\multicolumn{4}{l}{\bf LIDM Models + RL}\\
    &	LIDM, $I=50$,  +RL&	82.4	&	0.231\\
    &	LIDM, $I=70$,  +RL&	81.6	&	0.230\\
    &	LIDM, $I=100$, +RL&{\bf84.6}	&	0.240\\
    \bottomrule
  \end{tabular}}
  \caption{Corpus-based Evaluation.}
  \vspace{-5mm}
  \label{tab:lidm}
\end{table}

Table~\ref{tab:lidm} presents the results of the corpus-based evaluation.
The {\it Ground Truth} block shows the two metrics when we compute them on the human-authored responses.
This sets a gold standard for the task.
In the {\it Published Models} block, the results for the three baseline models were borrowed from Wen et al~\yrcite{wencond16}, they are: (1) the vanilla neural dialogue model (NDM), (2) NDM plus an attention mechanism on the belief trackers, and (3) the attentive NDM with self-supervised sub-task neurons.
The results of the LIDM model with and without RL fine-tuning are shown in the {\it LIDM Models} and the {\it LIDM Models + RL} blocks, respectively.
As can be seen, the initial policy learned by fitting the latent intention to the underlying data distribution  yielded reasonably good results on BLEU but did not perform well on task success when compared to their deterministic counterparts ({\it block 2 v.s. 3}).
This may be due to the fact that the variational lower bound of the dataset was optimised rather than task success during variational inference. 
However, once RL was applied to optimise the success rate as part of the reward function (Equation~\ref{eq:reward}) during the fine-tuning phase, the resulting LIDM+RL models  outperformed the three baselines in terms of task success without significantly sacrificing  BLEU ({\it block 2 v.s. 4})\footnote{Note that both NDM+Att+SS and LIDM use self-supervised information}.

In order to assess the human perceived performance, we evaluated the three models (1) NDM, (2) LIDM, and (3) LIDM+RL by recruiting paid subjects on Amazon Mechanical Turk.
Each judge was asked to follow a task and carried out a conversation with the machine.
At the end of each conversation the judges were asked to rate and compare the model's performance.
We assessed the subjective success rate,  the perceived comprehension ability and the naturalness of responses on a scale of 1 to 5.
For each model, we collected 200 dialogues and averaged the scores.
During human evaluation, we sampled from the top-5 intentions of the LIDM models and decoded a response based on the sample.
The result is shown in Table~\ref{tab:humaneavl}.
One interesting fact to note is that although the LIDM did not perform well on the corpus-based task success metric, the human judges rated its subjective success almost indistinguishably from the others.
This discrepancy between the two experiments arises mainly  from a flaw in the corpus-based success metric in that it favors greedy policies because the user side behaviours are fixed rather than interactional\footnote{The system tries to provide as much information as possible in the early turns, in case the fixed user side behaviours a few turns later do not fit  the scenario the system originally planned.}.
Despite the fact that LIDMs are considered only marginally better than NDM on subjective success, the LIDMs do outperform NDM on both comprehension and naturalness scores. 
This is because the proposed LIDM models can better capture multiple modes in the communicative intention and thereby respond more naturally by sampling from the latent intention variable.

\begin{table}[t]
  \centering
  \begin{tabular}{lccc}
    \toprule
    Metrics			&	NDM		&	LIDM	&	LIDM+RL\\
    \midrule
    Success			&	91.5\%	&	92.0\%	&	93.0\%\\
	Comprehension	&	4.21	&	4.40\tmark[{\makebox[0pt][l]{*}}]	&	4.40\\
    Naturalness		&	4.08	&	4.29\tmark[{\makebox[0pt][l]{*}}]	&	4.28\tmark[{\makebox[0pt][l]{*}}]\\
    \# of Turns		&	4.45	&	4.54	&	4.29\\
    \bottomrule
    \multicolumn{4}{l}{* p \textless 0.05}
  \end{tabular}
  \caption{Human evaluation. The significance test is based on a two-tailed student-t test, between NDM and LIDMs.}
  \vspace{-3mm}
  \label{tab:humaneavl}
\end{table}

\begin{table*}[t]
  	\centering
  	\Scale[0.85]{
  		\begin{tabular}{llll}
    	\toprule
Human :	& \multicolumn{3}{l}{Hi, I'm hungry for some Indonesian} \\
Machine:&{\bf( 0}&{\bf0.71)}& {\bf i am sorry , there are no indonesian restaurants in the area . would you like something else ?}\\
		&(46&0.19)& there are no restaurants serving indonesian kind of food .\\
        &(34&0.07)& what location would you like ?\\
                \midrule
Human :	& \multicolumn{3}{l}{Oh no. How about Indian?}\\
Machine:&( 0&0.59)& there are 9 restaurants in the east side of town . would you like their address ?\\
        &(30&0.15)& curry prince serves indian cuisine .\\
        &{\bf(34}&{\bf0.10)}& {\bf what area would you like ?}\\
        \midrule
Human : & \multicolumn{3}{l}{East side please}\\
Machine:&{\bf( 0}&{\bf0.62)}& {\bf curry prince is a indian restaurant in the east side of town . would you like their telephone number ?}\\
        &(84&0.19)& curry prince is an indian restaurant in the east of town and the number is 01223 566388 .\\
        &(81&0.08)& curry prince serves indian type of cuisine and is located in the east part of town .\\
        \midrule
Human : &\multicolumn{3}{l}{yes , the address and phone number please}\\
Machine:&( 3&0.46)& the address is 451 newmarket road fen ditton and the phone is 01223 566388 .\\
        &{\bf(11}&{\bf0.16)}& {\bf the address for curry prince is 451 newmarket road fen ditton and the phone number is 01223 566388 .}\\
        &\pbox{20cm}{( 0\\ }&\pbox{20cm}{0.15)\\ }& \pbox{20cm}{curry prince is located at 451 newmarket road fen ditton and their phone is 01223 566388 . is there \\anything else i can help you with ?}\\[7pt]
        \midrule
Human : &\multicolumn{3}{l}{Okay, thank you goodbye}\\
Machine:&( 0&0.57)& thank you , goodbye .\\
        &( 2&0.13)& goodbye .\\
        &{\bf( 1}&{\bf0.11)}& {\bf you are welcome . goodbye .}\\
        &( 6&0.10)& thank you for using the cambridge restaurant system . goodbye .\\
    \bottomrule
	\end{tabular}}
    \caption{A sample dialogue from the LIDM, I=100 model, one exchange per block. Each  latent intention is shown by a tuple (index, probability), followed by a decoded response. The sample dialogue was produced by following the responses highlighted in bold.}
    \label{tab:example}
    \vspace{-3mm}
\end{table*}

\begin{table*}[t]
  	\centering
  	\Scale[0.85]{
  		\begin{tabular}{llll}
    	\toprule
Human :	& \multicolumn{3}{l}{Hi, I'm hungry for some Indonesian .} \\
Machine:&{\bf( 0}&{\bf0.61)}& {\bf i am sorry , there are no indonesian restaurants in the area . would you like something else ?}\\
		&(46&0.17)& no restaurants in the part of town .\\
        &(34&0.14)& what part of town would you like ?\\
        \midrule
Human :	& \multicolumn{3}{l}{Oh no. How about Indian?}\\
Machine:&{\bf(93}&{\bf0.51)}& {\bf the gandhi is located at 72 regent street city centre and their telephone number is 01223 353942 .}\\
        &(72&0.21)& the gandhi serves indian kind of food and is in the centre of town and is in the cheap price range .\\
        &(12&0.12)& the gandhi is a indian restaurant in the centre location and is in the cheap price range .\\
        \midrule
Human : &\multicolumn{3}{l}{Okay, thank you goodbye}\\
Machine:&{\bf( 6}&{\bf0.28)}& {\bf thank you for using the cambridge restaurant system . goodbye .}\\
        &( 1&0.24)& you are welcome . goodbye .\\
        &( 85&0.19)& is there anything else i can help you with ?\\
    \bottomrule\toprule
Human : &\multicolumn{3}{l}{hey can i have some chinese cruisine in the centre ? and a cheap one please}\\
Machine:&(15& 0.42)& golden house is located at 12 lensfield road city centre and their number is 01842 753771 .\\
		&{\bf( 5}&{0.38\bf)}& {\bf golden house is located at 12 lensfield road city centre and the number is 01842 753771 . is there ...}\\
        &(93& 0.10)& golden house is located at 12 lensfield road city centre and their phone is 01842 753771 .\\
        \midrule
Human : &\multicolumn{3}{l}{that is all}\\
Machine:&{\bf( 6}&{\bf0.57)}& {\bf thank you for using the cambridge restaurant system . goodbye .}\\
        &( 1&0.20)& you are welcome . goodbye .\\
        &(17&0.11)& have a nice day !\\
    \bottomrule
	\end{tabular}}
    \caption{Two sample dialogues from the LIDM+RL, I=100 model, one exchange per block. Comparing to Table~\ref{tab:example}, the RL agent demonstrates a much greedier behavior toward task success. This can be seen in {\it block 2 \& block 4} in which the agent provides the address and phone number even before the user asks.}
    \label{tab:example2}
    \vspace{-3mm}
\end{table*}

Three example conversations are shown between a human judge and a machine, one from LIDM in Table~\ref{tab:example} and two from LIDM+RL in Table~\ref{tab:example2}, respectively.
The results are displayed one exchange per block. 
Each induced latent intention is shown by a tuple (index, probability)  followed by a decoded response, and the sample dialogues were produced by following the responses highlighted in bold.
As can be seen, the LIDM shown in Table~\ref{tab:example} clearly has multiple modes in the distribution over the learned intention latent variable, and what it represents can be easily interpreted by the response generated.
However, some intentions (such as {\it intent 0}) can result in very different responses under different dialogue states even though they were supervised by a small response set as shown in Table~\ref{tab:seed}.
This is mainly because of the variance introduced during variational inference.
Finally, when comparing Table~\ref{tab:example} and Table~\ref{tab:example2}, we can observe the difference between the two dialogue strategies: 
the LIDM, by inferring its policy from the supervised dataset, reflects the diverse set of modes in the underlying distribution; 
whereas the LIDM+RL, which refined its strategy using RL, exhibits a much greedier behavior in achieving task success (e.g. in \ {\it Table~\ref{tab:example2} in block 2 \& 4} the LIDM+RL agent provides the address and phone number even before the user asks).
This is also supported by the human evaluation in Table~\ref{tab:humaneavl} where LIDM+RL has much shorter dialogues on average  compared to the other two models.

\section{Discussion}\label{sec:discussion}

Learning an end-to-end dialogue system is appealing but challenging because of the credit assignment problem.
Discrete latent variable dialogue models such as LIDM are attractive because the latent variable can serve as an interface for decomposing the learning of language and the internal dialogue decision-making.
This decomposition can effectively help us resolve the credit assignment problem where different learning signals can be applied to different sub-modules to update the parameters.
In variational inference for discrete latent variables, the latent distribution is basically updated by the reward from the variational lower bound. 
While in reinforcement learning, the latent distribution (i.e. policy network) is updated by the rewards from dialogue success and sentence BLEU score.
Hence, the latent variable bridges the different learning paradigms such as Bayesian learning and reinforcement learning and brings them together under the same framework.
This framework provides a more robust neural network-based approach than previous approaches
because it does not depend solely on 
sequence-to-sequence learning
but instead explicitly models the hidden
 dialogue intentions underlying the user's utterances and allows the agent to directly learn a dialogue policy through interaction.  

\section{Related work}\label{sec:related}
Modeling chat-based dialogues~\cite{SerbanSBCP15,ShangLL15} as a sequence-to-sequence learning~\cite{SutskeverVL14} problem is a common theme in the deep learning community.
Vinyals and Le~\yrcite{VinyalsL15} has demonstrated a seq2seq-based model trained on a huge amount of conversation corpora which learns interesting replies conditioned on different user queries.
However, due to an inability to model dialogue context, these models generally suffer from the generic response problem~\cite{LiGBGD15,serban2016lv}.
Several approaches have been proposed to mitigate this issue, such as modeling the persona~\cite{liEtAl2016}, reinforcement learning~\cite{liEMNLP20162}, and introducing continuous latent variables~\cite{serban2016lv,cao17eacl}.
While in our case, we not only make use of the latent variable to inject stochasticity for generating natural and diverse machine responses but also model the hidden dialogue intentions explicitly.
This combines the merits of reinforcement learning and generative models. 

At the other end of the spectrum, goal-oriented dialogue systems typically adopt the POMDP framework~\cite{6407655} and break down the development of the dialogue systems into a pipeline of modules: natural language understanding~\cite{Henderson2015a}, dialogue management~\cite{6639297}, and natural language generation~\cite{wensclstm15}.
These system modules communicate through a dialogue act formalism~\cite{Traum1999}, which in effect constitute a fixed set of handcrafted intentions.  This limits the ability of such systems to scale to more complex tasks.
In contrast, the LIDM directly infers all underlying dialogue intentions from data and can handle intention distributions with long tails by measuring similarities against the existing ones during variational inference.
Modeling of end-to-end goal-oriented dialogue systems has also been studied recently~\cite{wencond16,wenN2N17,bordes16n2n}, however, these models are typically deterministic and rely on decoder supervision signals to fine-tune a large set of model parameters.

Much research has focused on combining different learning paradigms and signals to bootstrap performance.
For example, semi-supervised learning~\cite{NIPS2014_5352} has been applied in the sample-based neural variational inference framework as a way to reduce sample variance.
In practice, this relies on a discrete latent variable~\cite{miao16latentLangauage,kocisky16} as the vehicle for the supervision labels.
As in reinforcement learning, which has been a very common learning paradigm in dialogue systems~\cite{6639297,su2016acl,jiinloop2017}, the policy is also parameterised by a discrete set of actions.
As a consequence, the LIDM, which parameterises the intention space via a discrete latent variable, can automatically enjoy the benefit of bootstrapping from signals coming from different learning paradigms.
In addition, self-supervised learning~\cite{ilprints665} (or distant, weak supervision) as a simple way to generate automatic labels by heuristics is popular in many NLP tasks and has been applied to memory networks~\cite{hill2016} and neural dialogue systems~\cite{wencond16} recently.
Since there is no additional effort required in labeling, it can also be viewed as a method for bootstrapping.

\section{Conclusion}\label{sec:conclusion}

In this paper, we have proposed a framework for learning dialogue intentions via discrete latent variable models and introduced the Latent Intention Dialogue Model (LIDM) for goal-oriented dialogue modeling.
We have shown that the LIDM can discover an effective initial policy from the underlying data distribution and is capable of revising its strategy based on an external reward using reinforcement learning.
We believe this is a promising step forward for building autonomous dialogue agents since the learnt discrete latent variable interface enables the agent to perform learning using several differing paradigms.
The experiments showed that the proposed LIDM is able to communicate with human subjects and outperforms previous published results.

\section*{Acknowledgements}
Tsung-Hsien Wen is supported by Toshiba Research Europe Ltd, Cambridge Research Laboratory.
The authors would like to thank the members of the Cambridge Dialogue Systems Group for their valuable comments.



\bibliography{icml2017}
\bibliographystyle{icml2017}

\end{document}